\documentclass[final,5p,times,twocolumn]{elsarticle}

\usepackage{amssymb}
\usepackage{amsmath}
\usepackage{threeparttable}
\usepackage{tabularx}

\journal{}

\begin{document}

\begin{frontmatter}


\title{DMAGaze: Gaze Estimation Based on Feature Disentanglement and Multi-Scale Attention\tnotemark[*]}
\tnotetext[*]{The paper is under consideration at Pattern Recognition Letters} 

\author[label1]{Haohan Chen}
\affiliation[label1]{organization={Sichuan University},
            addressline={24 South Section 1, 1st Ring Road},
            city={Chengdu},
            postcode={610065},
            state={Sichuan Province},
            country={China}}
\author[label2]{Hongjia Liu}
\affiliation[label2]{organization={Aalto University},
            addressline={Maarintie 8, Espoo},
            city={Helsinki},
            postcode={02150},
            state={Uusimaa},
            country={Finland}}
\author[label1]{\corref{cor1} Shiyong Lan} 
\author[label3]{Wenwu Wang}
\affiliation[label3]{organization={University of Surrey},
            addressline={Stag Hill},
            city={Guildford},
            postcode={GU2 7XH},
            state={Surrey},
            country={UK}}
\author[label1]{Yixin Qiao}
\author[label1]{Yao Li}
\author[label1]{Guonan Deng}

\cortext[cor1]{Corresponding author: lanshiyong@scu.edu.cn.}

\begin{abstract}
Gaze estimation, which predicts gaze direction, commonly faces the challenge of interference from complex gaze-irrelevant information in face images. In this work, we propose DMAGaze, a novel gaze estimation framework that exploits information from facial images in three aspects: gaze-relevant global features (disentangled from facial image), local eye features (extracted from cropped eye patch), and head pose estimation features, to improve overall performance. Firstly, we design a new continuous mask-based Disentangler to accurately disentangle gaze-relevant and gaze-irrelevant information in facial images by achieving the dual-branch disentanglement goal through separately reconstructing the eye and non-eye regions. Furthermore, we introduce a new cascaded attention module named Multi-Scale Global Local Attention Module (MS-GLAM). Through a customized cascaded attention structure, it effectively focuses on global and local information at multiple scales, further enhancing the information from the Disentangler. Finally, the global gaze-relevant features disentangled by the upper face branch, combined with head pose and local eye features, are passed through the detection head for high-precision gaze estimation. Our proposed DMAGaze has been extensively validated on two mainstream public datasets, achieving state-of-the-art performance.
\end{abstract}




\end{frontmatter}

\section{Introduction}
Gaze estimation, the task of predicting gaze direction, crucial for measuring human attention, is widely applied in areas like saliency detection\cite{b1, b2}, virtual reality\cite{b3}, driver distraction monitoring\cite{b4}, human-computer interaction\cite{b5} and autism diagnosis\cite{b6}. Recently, gaze estimation has shifted from model-based methods to appearance-based methods. Model-based methods \cite{b15} rely on specialized sensors and 3D reconstruction techniques to capture the physical characteristics of the human eye and pupil, so as to estimate gaze direction. Whereas the appearance-based methods \cite{b21, b30, b7, b9, b18} are aimed at utilizing facial images and extracting image features through deep learning to achieve gaze estimation.

\begin{figure}[t]
    \centering
    \includegraphics[width=0.7\columnwidth]{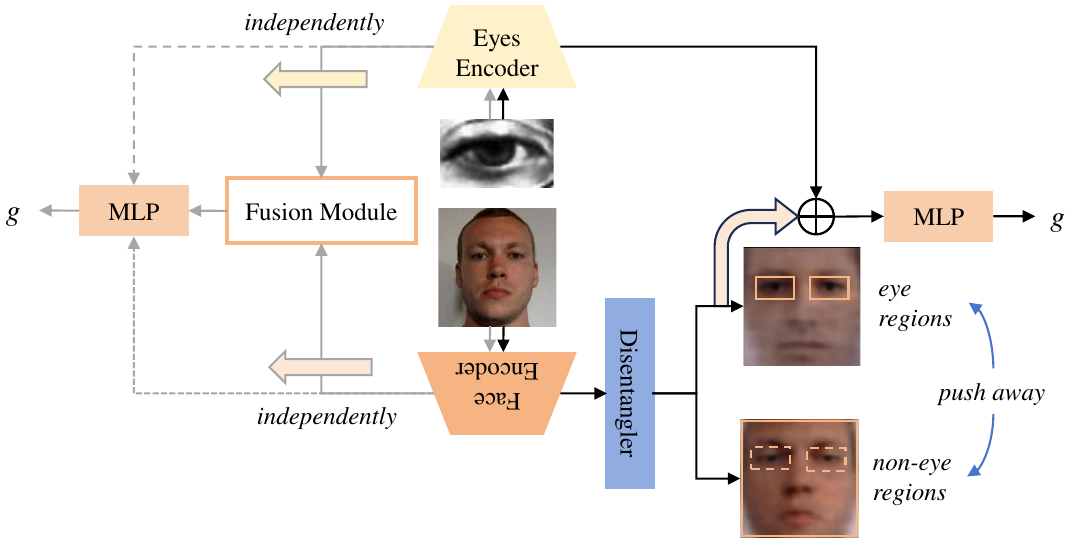} 
    \caption{This figure illustrates the comparison between traditional gaze estimation methods and our proposed method. The gray pathway represents the basic framework of many previous methods, while the black pathway represents our proposed framework for disentangling gaze-relevant and -irrelevant facial information to further explore the complex gaze relationship between eyes and face.}
    \label{fig1}
\end{figure}

In appearance-based methods, the input eye and facial images are used to estimate gaze either independently\cite{b30} \cite{b7} \cite{b9} \cite{b18} , in fusion\cite{b10} or through mutual assistance\cite{b8} \cite{b25}, as shown in the gray pathway on the left part of Fig.~\ref{fig1}. With only eye images as input, gaze estimation may be overly influenced by the eye’s state. Using only face images as input, gaze estimation can be degraded by varied information in them, such as facial appearances, expressions and poses.

\begin{figure*}[t]
    \centering
    \includegraphics[width=\textwidth]{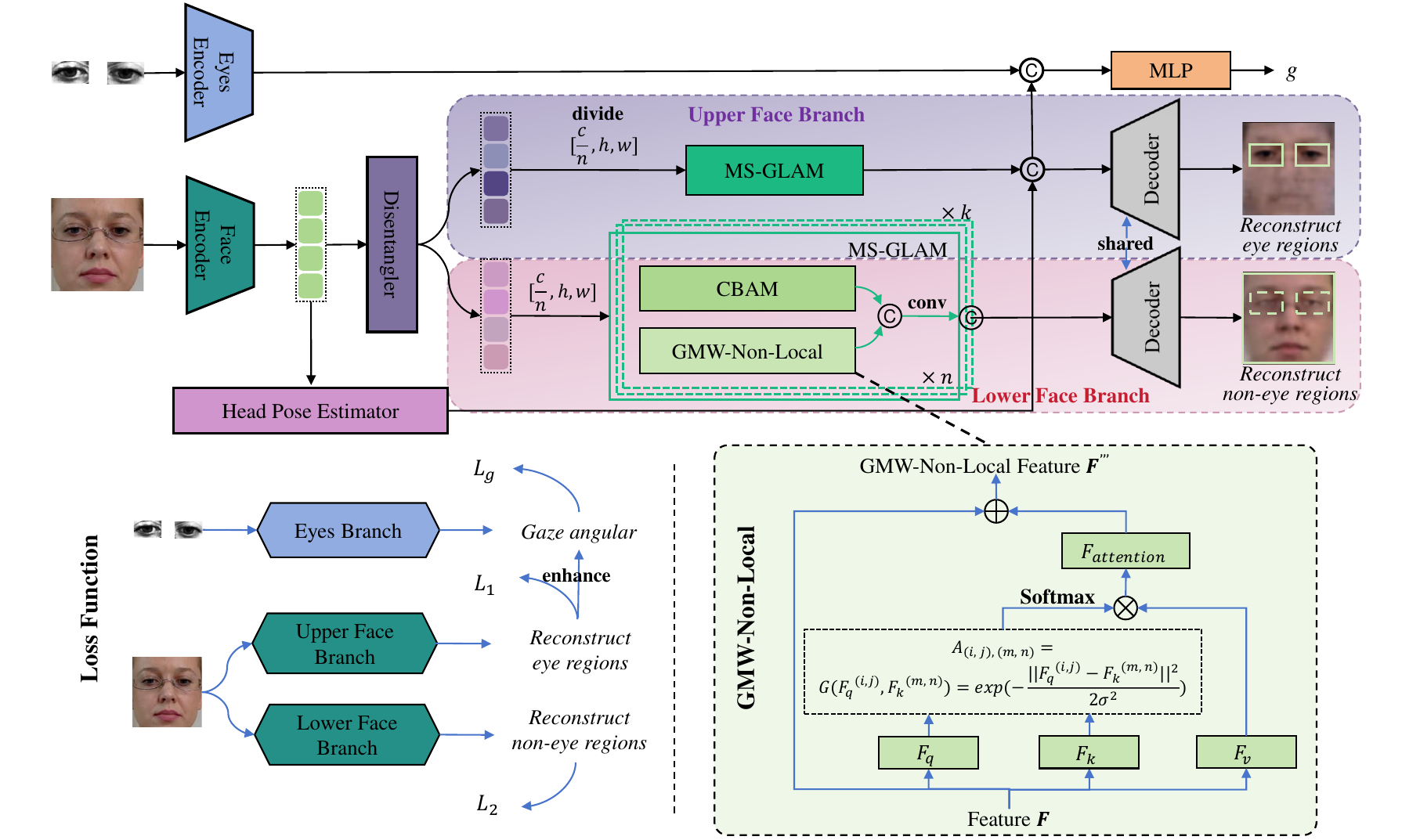} 
    \caption{The overall architecture of our proposed DMAGaze. The top illustrates the overall workflow of our model. The bottom left illustrates the propagation of the loss functions. The bottom right illustrates the GMW-Non-Local module.}
    \label{fig2}
\end{figure*}
  
Existing methods continue to evolve to address the above limitations, including using convolutions with spatial weights \cite{b7}, employing dilated convolutions \cite{b8} or coarse-to-fine adaptive networks \cite{b25} to extract useful features from the face and eyes, leveraging cross-attention to fuse features from both \cite{b10}, incorporating transformers into gaze estimation \cite{b9} and introducing facial text prompts \cite{b38} \cite{b11} inspired by the CLIP model \cite{b29} to guide gaze estimation. However, due to the highly intertwined nature of diverse facial information such as expressions, poses and appearances, it remains a challenge in disentangling gaze-relevant facial information for integrating global and local information for effective gaze estimation.

Convolutional Block Attention Module (CBAM)\cite{b13}\cite{b32} improve the capability of feature representation through channel and spatial weighting mechanisms. Non-local Neural Network (Non-Local)\cite{b14} captures global dependencies between features through non-local operators. However, as gaze-relevant and irrelevant information often entangles in high-dimensional space as nonlinear manifolds, the linear nature of matrix multiplication in Non-Local operators struggles to model complex nonlinear feature interactions. Therefore, the task of disentangling highly aggregated gaze-relevant and irrelevant information still warrants further exploration.

To address the above issues, we propose a Multi-Scale Global Local Attention Module (MS-GLAM) that connects CBAM and Non-Local through a custom cascaded attention structure to effectively integrate global and local information at multiple scales. Especially, via introducing a Gaussian Modulated Weighting (GMW), we incorporate Gaussian similarity-based distance metric into Non-Local, as shown in the bottom-right part of Fig.~\ref{fig2}. Instead of using traditional matrix multiplication for similarity computation, GMW utilizes the exponential decay of a distance metric, rather than directly relying on linear transformations of features, thereby stably modeling nonlinear global dependencies between the features. 

In addition, we propose DMAGaze, as shown in the top part of Fig.~\ref{fig2}, by exploiting the fact that the non-stationary \cite{b31} gazing process in gaze estimation datasets is primarily influenced by eye movement and head pose information, and that the facial information beneficial for eye reconstruction contains certain gaze-relevant details. The Disentangler, inspired by feature disentangling\cite{b12}, takes facial features as input and divides them into two sub-branches by individually focusing on the eye and non-eye regions of full face. By using continuous masks instead of binary ones in \cite{b12}, it can more flexibly capture the subtle differences between features, aiming to disentangle global gaze-relevant and gaze-irrelevant information. The distilled global gaze-relevant facial features effectively compensate for the limitations of the local information obtained solely from eye images for final gaze estimation. The above improvements enhance our model’s capability to estimate gaze in complex scenarios. Our main contributions are summarized as follows:

\begin{itemize}
\item We introduce a novel DMAGaze framework. By using the proposed Disentangler to disentangle global gaze-relevant and gaze-irrelevant facial features from facial images, we combine the global gaze-relevant facial information with our estimated head pose and local eye information, ultimately improving gaze estimation.
\item We propose a MS-GLAM with a customized cascaded attention structure to effectively enhance global and local information. Additionally, we introduce a GMW into Non-Local, enhancing non-linear modeling capability through Gaussian similarity-based distance metrics.
\item Finally, we conduct extensive comparisons on mainstream datasets, demonstrating the superiority of our gaze estimation method.
\end{itemize}

\section{Related work}
\subsection{Gaze estimation}

The development of gaze estimation is closely linked to technological advancements. It can be mainly divided into traditional model-based methods that use eye geometric models and infrared scanning technology, and the current mainstream appearance-based methods that have emerged with the development of deep learning. Model-based methods\cite{b15} often rely on dedicated hardware and sensors to collect data and build models based on the geometric features of the eyes. Data collection in these methods usually requires users to wear specific devices and perform designated actions multiple times. These methods are extremely sensitive to individual differences and changes in environmental conditions. Limited by equipment, personnel, and the environment, such gaze estimation methods have gradually faded from the spotlight. Appearance-based methods\cite{b30, b7, b9, b18} are deeply influenced by deep learning. They typically take facial or eye images as input, extract features through neural networks, and then perform gaze estimation.

Zhang et al.\cite{b21, b7} were the first to introduce CNN into the gaze estimation task and later used the attention mechanism to weight facial features. Fisher et al.\cite{b17} used two VGG networks to process two eye images. Cheng et al.\cite{b25} first estimated gaze from facial images and then refined it using eye images. Krafka\cite{b40} et al. proposed a multi-channel network that takes eye images, full-face images, and facial mesh information as inputs. Cheng et al.\cite{b28} used generative adversarial networks to purify gaze-related features in facial images. Abdelrahman et al.\cite{b39} used pitch and yaw dual branches for cross-entropy and regression iterations on facial image features for gaze estimation. Cătrună et al.\cite{b10} fused facial and eye features through cross-attention for gaze estimation.

It can be observed that there are various combinations of inputs for the gaze estimation problem, such as eye images, facial images, and even head pose information. Eye images carry crucial information about the line of sight, and \cite{b40, b8, b25} have verified the effectiveness of facial images in improving the performance of this task. This paper follows this conclusion and uses both facial and eye images as inputs to the network. However, current fusion methods for different inputs are mostly simple concatenations or rely on attention mechanisms, making it difficult to eliminate the influence of noise information unrelated to gaze. Moreover, there is still a lack of in-depth exploration of the different contributions of local and global information from the eyes and face to gaze estimation.

\subsection{Gaussian function}
The Gaussian distribution is a distribution function that describes the probability characteristics of random variables and is commonly used in statistics to represent the distribution of data. The Gaussian kernel is constructed based on the probability density function of the Gaussian distribution and uses the exponential form of the Gaussian distribution to measure the similarity between data points. Its concept can be traced back to the fields of statistics and signal processing, especially in Gaussian processes and kernel methods. In the early 1990s, the Gaussian kernel began to be systematically applied in machine learning and computer vision\cite{b41}. As a landmark algorithm in this field, the Support Vector Machine (SVM) often uses kernel functions to handle non-linear data, and the Gaussian kernel is one of the commonly used kernel functions.

Due to its smoothness and robustness, the Gaussian kernel is widely applied in various fields such as image denoising, edge detection, and image segmentation. For example, in image segmentation, the Gaussian kernel is used to measure similarity to optimize the segmentation process\cite{b42}. In image denoising and blurring, the Gaussian kernel is used to approximate the calculation of the bilateral filter\cite{b43}. In classification tasks, the deep Gaussian process is applied to enhance the robustness of the model\cite{b44}. In the reconstruction of complex scenes of Neural Radiance Fields (NeRF), the Gaussian kernel is used to reduce noise and improve detail preservation\cite{b45}.

In recent years, researchers have started to use the Gaussian kernel for feature matching and similarity calculation\cite{b46, b47, b48}, especially in tasks that require detailed modeling of local features. These applications have gradually confirmed the characteristic of the Gaussian kernel that it can smooth noise while maintaining important structural features. This also highlights the urgent problem to be solved by the model in this paper: during the process of reconstructing the eye and non-eye regions with the full-face dual-branch method, there are problems of unstable attention mechanisms and noises between different regions when decoupling and extracting multi-scale features. Therefore, we propose to use the Gaussian similarity-based distance metric in the attention module to replace the traditional dot-product similarity calculation method. The method converts the Euclidean distance between feature points into a probability measure, so that feature points closer in distance are assigned higher weights, while those farther apart are assigned lower weights. This improvement can more effectively smooth local details and reduce the sensitivity to gaze-irrelevant features compared with the traditional method. It not only expands the application scope of the attention module but also provides a smoother and more robust way for feature extraction.

\section{Methodology}
A high-level overview of our proposed DMAGaze structure is shown in Fig.~\ref{fig2}. Eye images are processed through an eye encoder in the eye branch to extract eye features, while facial images pass through a face encoder in the face branch. Gaze-relevant features are then distilled through the Disentangler and MS-GLAM by reconstructing the eye and non-eye regions of the full-face images using the decoders. The distilled gaze-relevant features are subsequently combined with head pose features extracted by simple convolutions and local eye features to predict gaze.

We input left and right eye images $I^l$ and $I^r$ of size $H^e \times W^e \times 3$ into the eyes encoder $\text{Enc}^e$, and the facial image $I$ of size $H^f \times W^f \times 3$ into the face encoder $\text{Enc}^f$. The eyes encoder extracts the eye features $F^e \in \mathbb{R}^{n \times d}$ ($n$ is the number of features, $d$ is the dimension of features), while the face encoder extracts the initial full-face features $F^f \in \mathbb{R}^{c \times h \times w}$ ($c$ is the number of channels, $h$ is the height and $w$ is the width). This process can be formalized as:
\begin{equation}
    F^e = \text{Enc}^e(I^l, I^r) \label{eq1}
\end{equation}
\vspace{-15pt}
\begin{equation}
    F^f = \text{Enc}^f(I) \label{eq2}
\end{equation}

\subsection{Disentangler} The initial facial features $F^f$ extracted in the facial branch are disentangled by the Disentangler into two branches, which are used to further extract gaze-relevant features $F^r$ and gaze-irrelevant features $F^{ir}$. $F^r$, as the global gaze-relevant features extracted from the facial images, compensate for the limitations of directly using the local eye images. This process is defined as follows:
\begin{equation}
    F^r = k \odot F^f \label{eq3}
\end{equation}
\vspace{-15pt}
\begin{equation}
    F^{ir} = (1-k) \odot F^f \label{eq4}
\end{equation}
Where $k \in [0,1]^{c \times h \times w}$ and $1-k \in [0,1]^{c \times h \times w}$ serve as learnable multi-variable weight matrices with the same shape as the facial features $F^f$ controlling the selective transmission of feature information, and $\odot$ denotes element-wise multiplication.

\subsection{Multi-scale global-local attention module} The disentangled dual-branch facial features $F^r$ and $F^{ir}$ are both input into the MS-GLAM. To deeply disentangle gaze-relevant and gaze-irrelevant features, we introduce GMW into Non-Local\cite{b14}, using Gaussian similarity-based distance metric to replace dot product calculation, indirectly capture non-linear relationships in a implicit higher-dimensional space, enhancing the ability to capture global feature dependencies, as shown in Fig.~\ref{fig2}. The Gaussian similarity used in GMW, as a Gaussian distribution-based function, has its similarity measurement defined as follows:
\begin{equation}
    G(q, k) = \text{exp}(-\frac{{||q - k||}^2}{2\sigma^2}) \label{eq5}
\end{equation}
Where $q$ and $k$ are the input feature vectors, and $\sigma$ is the variance parameter.

\paragraph{Convolutional block attention module} CBAM combines channel and spatial attention mechanisms for feature selection. Assuming the input feature is $F \in \mathbb{R}^{c \times h \times w}$, the entire process can be expressed as follows:
\begin{equation}
    F^{'} = \text{M}_c(F) \odot F \label{eq6}
\end{equation}
\vspace{-15pt}
\begin{equation}
    F^{''} = \text{M}_s(F^{'}) \odot F^{'} \label{eq7}
\end{equation}
Where $\text{M}_c$ and $\text{M}_s$ are the channel attention and spatial attention operations respectively. $F^{'}$ and $F^{''}$ represent the channel-refined feature and CBAM feature respectively. The channel attention module includes average pooling and max pooling layers along the spatial dimension:
\begin{equation}
    F^c_{avg} = \text{avgpooling}(F)\label{eq8}
\end{equation}
\vspace{-15pt}
\begin{equation}
    F^c_{max} = \text{maxpooling}(F)\label{eq9}
\end{equation}
\vspace{-15pt}
\begin{equation}
    F_{c} = \text{sigmoid}(\text{MLP}(F^c_{avg}) + \text{MLP}(F^c_{max}))\label{eq10}
\end{equation}
Where $\text{sigmoid}(\cdot)$ is a sigmoid activation function,  $\text{MLP}(\cdot)$ is a multi-layer perceptron, and $F_c \in \mathbb{R}^{c \times 1 \times 1}$ represents the channel attention map. Then the average and maximum values are taken along the channel dimension. The initial spatial attention map output from the convolution layer is used to produce a spatial attention map as follows:
\begin{equation}
    F^s_{avg} = \frac{1}{C}\sum_{c=1}^C{F^{'}[:, C, :, :]} \label{eq11}
\end{equation}
\vspace{-15pt}
\begin{equation}
    F^s_{max} = \text{max}_{c=1...C}F^{'}[:, C, :, :] \label{eq12}
\end{equation}
\vspace{-15pt}
\begin{equation}
    F_s = \text{sigmoid}(\text{conv}(\text{cat}(F^s_{avg}, F^s_{max}))) \label{eq13}
\end{equation}
Where $\text{conv}(\cdot)$ is the convolution operation, $\text{cat}(\cdot)$ is the feature concatenation operation and $F_s \in \mathbb{R}^{1 \times h \times w}$ represents the output spatial attention map.

\paragraph{Gaussian modulated weighting-non-local} Non-Local utilizes non-local operators to capture long-distance feature dependencies. However, its similarity measurement based on dot product computation struggles to fully capture complex nonlinear relationships. To address this, we introduce the GMW, as shown in the bottom-right part of Fig.~\ref{fig2}. The exponential form of the Gaussian similarity formula enables Gaussian Modulated Weighting-Non-Local (GMW-Non-Local) to perform nonlinear similarity measurement based on distances, indirectly capturing relationships in a implicit higher-dimensional space. This enhances our model's ability to model global feature hierarchies. We take the input feature $F \in \mathbb{R}^{c \times h \times w}$ and apply convolutional layers to transform it:
\begin{equation}
    F_q = \text{conv}_q(F) \label{eq14}
\end{equation}
\vspace{-15pt}
\begin{equation}
    F_k = \text{conv}_k(F) \label{eq15}
\end{equation}
\vspace{-15pt}
\begin{equation}
    F_v = \text{conv}_v(F) \label{eq16}
\end{equation}
In this case, $\text{conv}_q(\cdot)$, $\text{conv}_k(\cdot)$ and $\text{conv}_v(\cdot)$ represent the convolutional transformations. The attention weight matrix $A \in \mathbb{R} ^{(h_q \times w_q) \times (h_k \times w_k)}$ between the query $F_q$ and the key $F_k$ is defined as:
\begin{equation}
\begin{aligned}
     A_{(i, j), (m, n)} &= G(F_q^{(i,j)}, F_k^{(m,n)})\\
    &= \text{exp}(-\frac{||F_q^{(i,j)} - F_k^{(m,n)}||^2}{2\sigma^2}) \label{eq17}
\end{aligned}
\end{equation}
Where $h_q, w_q$ and $h_k, w_k$ are the heights and widths of $F_q$ and $F_k$, respectively. $i, j$ and $m,n$ denote the spatial position indices of $F_q$ and $F_k$, respectively. $F_q^{(i,j)}$ and $F_k^{(m,n)}$ are the input feature vectors of GMW. The final output $F^{'''} \in \mathbb{R}^{c \times h \times w}$ is given by:
\begin{equation}
    F^{'''} = \text{softmax}(A) \times F_v + F \label{eq18}
\end{equation}
where $\text{softmax}(\cdot)$ is the activation function.

\paragraph{Cascaded attention structure}
We design a cascaded structure for the above two attention modules, which not only integrates local and global information effectively but also extends the multi-scale capturing capability. We take the input feature $X \in \mathbb{R}^{c \times h \times w}$ and divide it into $n$ groups along the channel dimension. $X_i \in \mathbb{R}^{\frac{c}{n} \times h \times w}$ represents the grouped feature and $i \in \{0, 1, ..., n-1\}$ represents the $i$-th group. After applying convolutional transformation to each group of features, the CBAM and GMW-Non-Local operations are applied. This process is repeated for each group until all groups have been processed. The entire process can be formalized as:
\begin{equation}
    x_{sub}^i = \text{conv}_{\text{sub}}(X_i) \label{eq21}
\end{equation}
\begin{equation}
a_i = 
\left\{
\begin{aligned}
    & \text{CBAM}(x_{sub}^i), & i = 0 \\
    & \text{CBAM}\left(\text{cat}(z_{i-1}, x_{sub}^i)\right), & 1 \leq i \leq n-1
\end{aligned}
\right.
\end{equation}
\begin{equation}
b_i = 
\left\{
    \begin{aligned}
      & \text{GMW}(x_{sub}^i), & i = 0 \\
      & \text{GMW}\left(\text{cat}(z_{i-1}, x_{sub}^i)\right), & 1 \leq i \leq n-1
    \end{aligned}
\right.
\end{equation}
\begin{equation}
    z_i = \text{conv}_{\text{tail}}(\text{cat}(a_i, b_i)) \label{eq22} 
\end{equation}
\begin{equation}
    Z = \text{cat}(z_0, z_1, ..., z_{n-1}) \label{eq23}
\end{equation}
Where $\text{conv}_{\text{sub}}(\cdot)$ represents the convolutional operation for subgroups, $\text{CBAM}(\cdot)$ and $\text{GMW}(\cdot)$ represent the CBAM operation and GMW-Non-Local operation respectively, and $\text{conv}_{\text{tail}}(\cdot)$ represents the convolutional operation at the end. Traversing all subgroups constitutes one round. We choose to carry out $k$ rounds. The initial input for each round is as follows:
\begin{equation}
X^j = 
\left\{
    \begin{aligned}
      &X, & j=0 \\
      & \text{conv}(\text{cat}(Z^{j-1}, X^{j-1}), & 1\leq j \leq k-1
    \end{aligned}
\right.
\end{equation}
Where $X^j$ is the initial input before the $j$-th round of grouping, $Z^{j-1}$ is the output after the $(j-1)$-th round of MS-GLAM.

\subsection{Loss function} The facial dual-branch features $F^r$ and $F^{ir}$ are processed through MS-GLAM and then decoded with the aim of disentangling gaze-relevant and gaze-irrelevant facial features. Based on the cognitive premise that facial features beneficial for eye region reconstruction inherently contain gaze-relevant information, the upper and lower facial branches further disentangle facial information by reconstructing the eye and non-eye regions of the full-face images. The propagation of the loss function in our overall framework is shown in the bottom-left part of Fig.~\ref{fig2}. The reconstruction loss for the eye regions $L_1$ is defined as follows:
\begin{equation}
    L_1 = \text{MSE}(\hat{I^l}, I^l) + \text{MSE}(\hat{I^r}, I^r) \label{eq24}
\end{equation}
Here, $\hat{I^l}$ and $\hat{I^r}$ represent the reconstructed images of the left and right eyes respectively, while $I^l$ and $I^r$ denote the corresponding original images. $\text{MSE}(\cdot)$ represents the mean squared error. The reconstruction loss for the non-eye regions $L_2$ is defined as follows:
\begin{equation}
    L_2 = \text{MSE}(\hat{top}, top) + \text{MSE}(\hat{mid}, mid) + \text{MSE}(\hat{bot}, bot) \label{eq25}
\end{equation}
Here, $\hat{top}$, $\hat{mid}$ and $\hat{bot}$ correspond to the reconstructed top, middle and bottom regions of the full face with the eye regions removed, while $top$, $mid$ and $bot$ denote their respective original regions.

The distilled facial upper-branch features are combined with local eye features and head pose features to jointly predict gaze angles. The loss of gaze estimation $L_g$, which measures the discrepancy between the predicted gaze direction and the ground truth, is defined as:
\begin{equation}
    L_g = \frac{1}{N}\sum_{i=1}^N{|\hat{g_i}-g_i|} \label{eq26}
\end{equation}
Here, $N$ is the number of samples, $\hat{g_i}$ and $g_i$ represent the predicted and ground truth gaze angles for the $i$-th sample respectively, and $|\cdot|$ indicates the element-wise absolute value operation.

\begin{figure}[t]
    \centering
    \includegraphics[width=0.8\columnwidth]{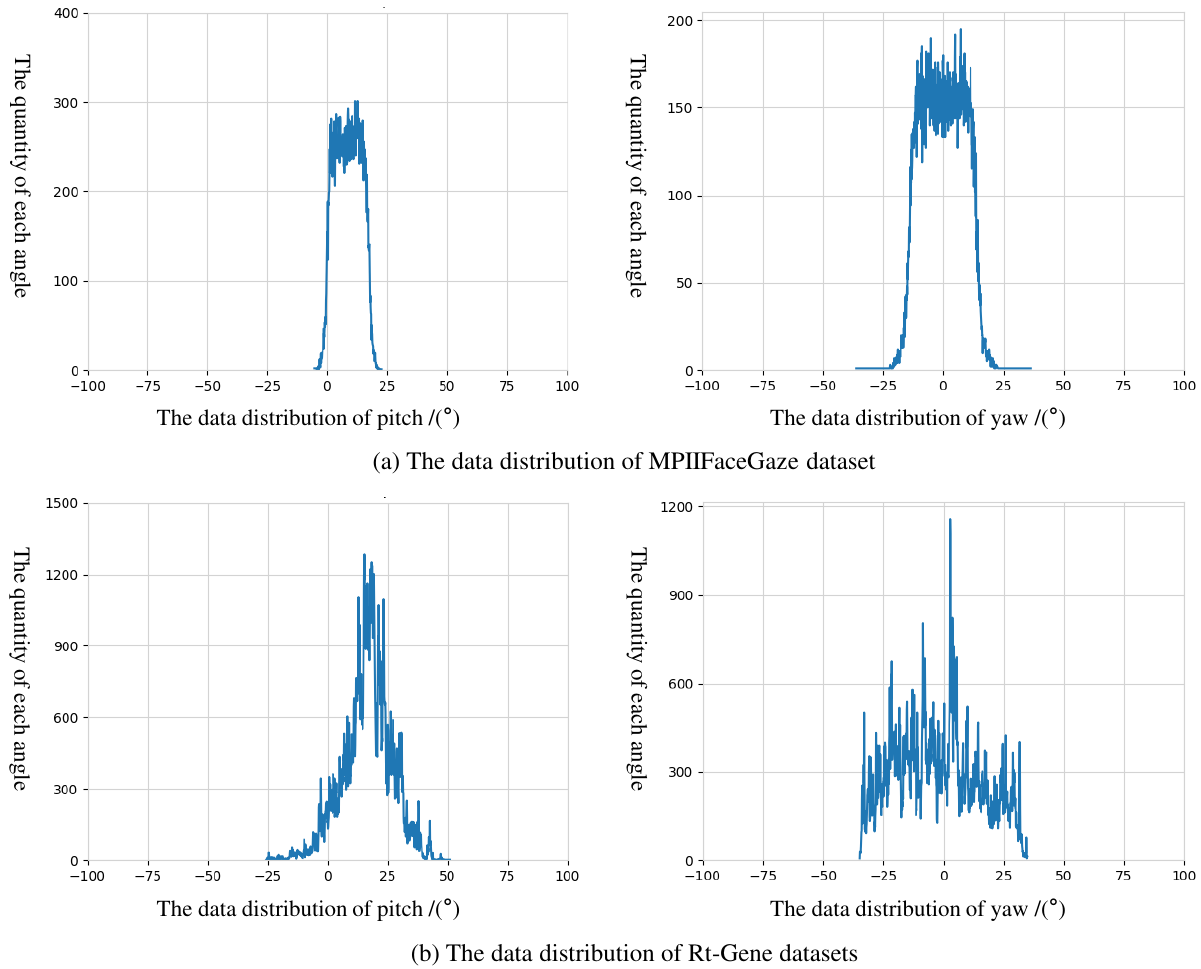} 
    \caption{The data distribution of MPIIFaceGaze and Rt-Gene dataset.}
    \label{fig5}
\end{figure}

\section{Experiments}
\subsection{Datasets}
To assess the performance of our model, experiments are conducted on publicly available gaze estimation datasets: MPIIFaceGaze\cite{b7} and Rt-Gene\cite{b17}. The MPIIFaceGaze dataset is smaller in scale, while the Rt-Gene dataset is larger. All datasets are preprocessed as described in \cite{b20} for fairness, and eye images are cropped using eye corner position information. To compute the reconstruction loss, the eye corner position information in MPIIFaceGaze and Rt-Gene is added to the label files for model training.

\textbf{MPIIFaceGaze.} The MPIIFaceGaze dataset is an extension of the MPIIGaze\cite{b21} dataset, providing 45K facial images captured over several months from 15 subjects using a laptop camera. It includes various lighting conditions, facial expressions and head poses in an unconstrained environment.

\textbf{Rt-Gene.} The Rt-Gene dataset contains 122K images from 15 subjects wearing eye-tracking glasses (with two subjects recorded twice). The subjects were positioned between 0.5 and 2.9 meters from the camera, with notable variations in head pose.

\subsection{Setup}
\textbf{Train.} Our experiments are implemented with PyTorch and conducted on NVIDIA GeForce RTX 4090. We train DMAGaze with a batch size of 48 and for 40 epochs on the MPIIFaceGaze datasets, and with a batch size of 64 and for 40 epochs on the Rt-Gene dataset. The AdamW optimizer \cite{b23} is used for model optimization, with $\beta_1 = 0.9$ and $\beta_2 = 0.999$. We adjust the learning rate using MultiStepLR \cite{b23} with an initial learning rate of $1e-4$, milestones at $[10, 25]$ and a learning rate decay factor of $0.1$.

\textbf{Implementation details.} The input consists of normalized face images of size $224 \times 224 \times 3$ and eye images of size $36 \times 60 \times 3$. The face encoder in DMAGaze utilizes a pre-trained ResNet34. In MS-GLAM, the number of rounds $k$ for the cascaded attention structure is set to $4$, and the number of the input feature groups $n$ is set to $4$. The variance parameter $\sigma$ in GMW is set to $1.0$. We use the most widely used evaluation metric in gaze estimation, the angular error, which measures the angular difference between the predicted and ground truth 3D gaze vector. A smaller angular error indicates better gaze estimation performance. It's defined as:

\begin{equation}
    \text{Angular Error} = \text{arccos}(\frac{g \cdot g^*}{||g|| \cdot ||g^*||}) \label{eq27}
\end{equation}

\begin{table*}[t]
\caption{Comparison with the state-of-the-art methods} \label{tab1}
    \centering   
        \begin{tabular}{l l l l l}
            \hline
            Methods & Year & MPIIFaceGaze & Rt-Gene & Inputs \\
            \hline
            FullFace\cite{b7} & 2017 & 4.93° & 10.00° & face ($448 \times 448$) \\ 
            RT-GENE\cite{b17} & 2018 & 4.66° & 8.60° & eye \\
            Dilated-Net\cite{b8} & 2018 & 4.42° & 8.38° & eye ($64 \times 96$) and face ($96 \times 96$) \\
            Gaze360\cite{b18} & 2019 & 4.06° & 7.06° & face \\
            CA-Net\cite{b25} & 2020 & 4.27° & 8.27° & eye ($60 \times 36$) and face ($224 \times 224 \times 3$)  \\
            AGE-Net\cite{b26} & 2021 & 4.09° & 7.44° & eye ($60 \times 36$) and face ($120 \times 120$) \\
            GazeTR\cite{b9} & 2022 & 4.00° & 6.55° & face ($224 \times 224 \times 3$) \\
            GazeCaps\cite{b33} & 2023 & 4.06° & 6.92° & face ($224 \times 224 \times 3$) \\
            EM-Net\cite{b36} & 2024 & 3.88° & 6.27° & face ($224 \times 224 \times 3$) \\ 
            GazeSetMerge\cite{b37} & 2024 & 3.88° & 6.46° & eye ($128 \times 128 \times 3$) and face ($224 \times 224 \times 3$) \\
            DMAGaze (Ours) & 2025 & \textbf{3.74°} & \textbf{6.17°} & eye($60 \times 36$) and face ($224 \times 224 \times 3$) \\
            \hline
        \end{tabular}
\end{table*}

\subsection{Comparison with the state-of-the-arts}
\begin{figure}[t]
    \centering
    \includegraphics[width=0.7\columnwidth]{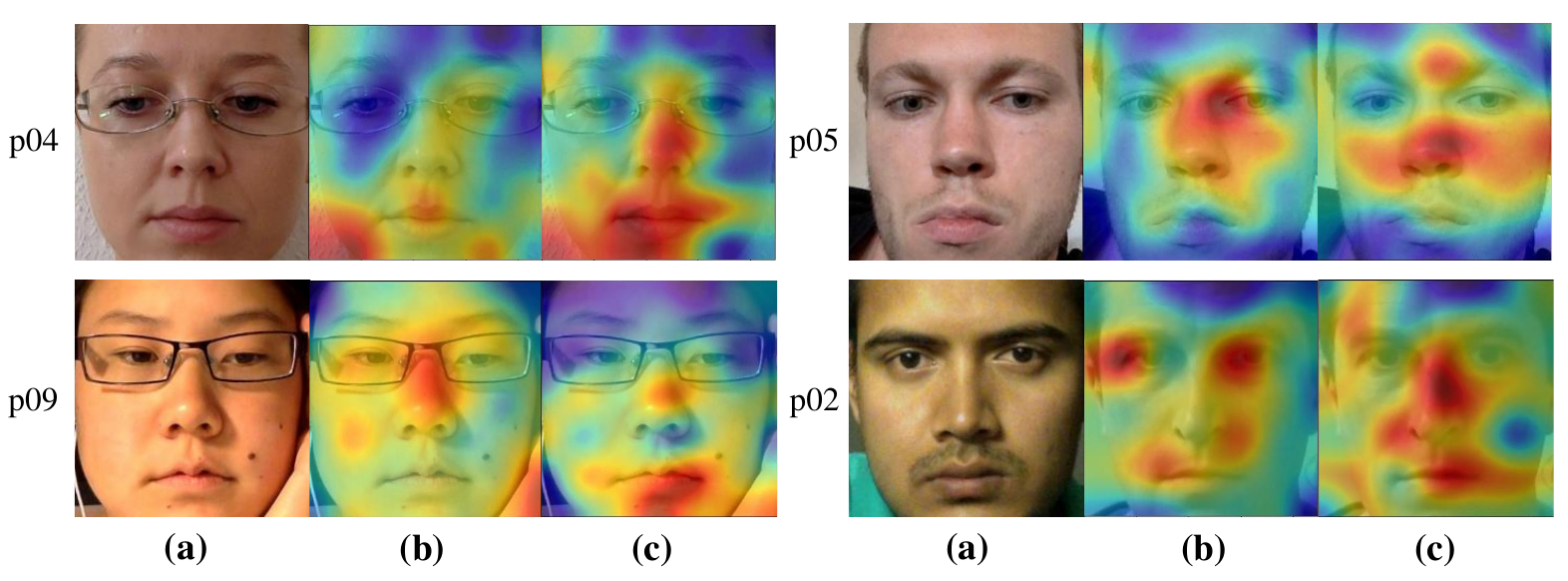} 
    \caption{The visualization of the attention maps of the upper face branch and lower face branch of DMAGaze we proposed. (a) Input images from MPIIFaceGaze dataset. (b) Attention maps from the upper face branch after Disentangler, which is mainly responsible for reconstructing the eye region. (c) Attention maps from the lower face branch after Disentangler, which is dedicated to reconstructing non-eye regions.}
    \label{fig4}
\end{figure}

We compare DMAGaze with state-of-the-art methods on the MPIIFaceGaze and Rt-Gene datasets, which are widely recognized for their complexity and diversity in gaze estimation tasks. From the analysis of the data distribution characteristics in Fig.~\ref{fig5}, the sample points of the MPIIFaceGaze and Rt-Gene datasets are mainly concentrated in the angular intervals of $\pm$25° and $\pm$50°, exhibiting a significant distribution imbalance. However, this inherent challenge in data distribution precisely provides critical support for validating the effectiveness of our proposed method on gaze estimation. As shown in Table~\ref{tab1}, our method achieves angular errors of 3.74° and 6.17° on these datasets. It is evident that our model outperforms all others, achieving performance improvements of $3.61\%$ and $1.59\%$, respectively. This result demonstrates the effectiveness of the DMAGaze framework we proposed, which disentangles gaze-relevant and gaze-irrelevant facial information using the Disentangler and MS-GLAM, and integrates global gaze-relevant features with head pose and local eye features for accurate gaze estimation. Notably, FullFace, Gaze360, GazeTR, GazeCaps and EM-Net rely solely on facial images for gaze estimation, whereas Rt-Gene uses only eye images. In contrast, others methods, including ours, use both eye and facial images. The visualization of attention maps for the upper and lower facial branches in Fig.~\ref{fig4} further validates our proposed framework. The upper branch, mainly for reconstructing eye regions, focuses on eyes with non-eye regions providing some assistance. The lower branch, tasked with non-eye regions reconstruction, concentrates on those areas such as the nose, mouth, cheeks and contours. These results show the distinct focus areas of the two branches and confirm the framework's effectiveness in separating gaze-relevant and gaze-irrelevant information.

\begin{table*}[t]
    \caption{Comparative experiments on MPIIFaceGaze dataset}
    \begin{center}
    \begin{threeparttable} 
    \begin{tabular}{l l l l}
    \hline
    Trial number & Multi-scale & Attention module selection & Angular error \\
    \hline
    \multicolumn{4}{l}{\textit{Independent replacement}} \\
    \hline
     1 & \checkmark & CBAM & 3.82° \\
     2 & \checkmark & GAM & 3.86° \\
     3 & \checkmark & SCSA & 3.83° \\
     4 & \checkmark & Non-Local & 3.85° \\
     5 & \checkmark & GMW-Non-Local & \textbf{3.81°} \\
     6 & \checkmark & Agent Attention & 3.84° \\
    \hline
    \multicolumn{4}{l}{\textit{Simple concatenation}} \\
    \hline
     7 & × & CBAM & 3.85° \\
     8 & × & GAM & 3.89° \\
     9 & × & SCSA & 3.85° \\
     10 & × & GMW-Non-Local & \textbf{3.83°} \\
     11 & × & Agent Attention & 3.91° \\
    \hline
    \multicolumn{4}{l}{\textit{Combined configuration}} \\
    \hline
     12 & \checkmark & CBAM + GMW-Non-Local & \textbf{3.74°} \\
     13 & \checkmark & GAM + GMW-Non-Local & 3.82°\\
     14 & \checkmark & SCSA + GMW-Non-Local & 3.83° \\
     15 & \checkmark & CBAM + Agent Attention & 3.83° \\
     16 & \checkmark & GAM + Agent Attention & 3.87° \\
     17 & \checkmark & SCSA + Agent Attention & 3.82° \\
    \hline
    \end{tabular}
    \begin{tablenotes} 
        \item * \textbf{Multi-scale} stands for whether the attention module is applied to our proposed cascaded attention structure.
    \end{tablenotes}
    \end{threeparttable}
    \label{tab5}
    \end{center}
\end{table*}

\begin{table}[t]
\caption{Ablation study of DMAGaze}
    \begin{center}
        \begin{tabular}{ll}
        \hline
        Method & Angular error \\
        \hline
        Baseline & 5.02° \\
        + Face branch (w/o MS-GLAM) & 4.04° \\
        + Head pose & 3.86° \\
        + MS-GLAM (w/o GMW) & 3.79° \\
        + GMW (total model) & \textbf{3.74°} \\
        \hline
        \end{tabular}
        \label{tab3}
    \end{center}
\end{table}

\subsection{Selection of Attention Modules}
To validate the effectiveness of attention module selection and combinatorial strategies in our proposed MS-GLAM framework, we conduct systematic comparative evaluations on the MPIIFaceGaze dataset. The experiments focus on two critical feature enhancement dimensions: hybrid channel-spatial attention and global contextual attention, as detailed in Table~\ref{tab5}. For hybrid channel-spatial attention mechanisms, we select representative methods covering distinct computational paradigms, including CBAM\cite{b32} (sequential channel-spatial refinement), GAM\cite{b49}(parallel 3D attention weighting), and SCSA\cite{b50}(which explores synergistic effects between spatial and channel attention via a Shareable Multi-Semantic Spatial Attention module and Progressive Channel Self-Attention module). For global dependency modeling, we compare representative long-range context modeling methods covering Non-Local\cite{b14}(standard self-attention), Agent Attention\cite{b51}(which introduces proxy tokens to aggregate and broadcast global information, combining the advantages of Softmax and Linear Attention), and our improved Non-Local enhanced with Gaussian Modulated Weighting (GMW). To ensure fair comparison, all attention modules are integrated into the identical network position within MS-GLAM while maintaining identical input/output dimensions, and the number of iterations $k$ is set to 4 for all cases.

\begin{table}[t]
\caption{Ablation study on parameter $k$ and $\sigma$}
    \begin{center}
        \begin{tabular}{l l l l}
        \hline
        Rounds $k$ & Variance parameter $\sigma$ & Angular error \\
        \hline
        2 & 1.0 & 3.87° \\
        4 & 1.0 & \textbf{3.74°} \\
        8 & 1.0 & 3.81° \\
        \hline
        4 & 0.5 & 3.87° \\
        4 & 1.0 & \textbf{3.74°} \\
        4 & 2.0 & 3.84° \\
        \hline
        \end{tabular}
        \label{tab4}
    \end{center}
\end{table}

We investigate three configurations: independent replacement, simple concatenation, and combined configuration. The independent replacement section verifies the independent impact of each attention module on gaze estimation by individually embedding different attention mechanisms into the cascaded attention framework for comparison. As shown in Table~\ref{tab5}, the GMW-Non-Local module outperforms the standard Non-Local module with a 1.04\% improvement. This result not only explains the reason for selecting GMW-Non-Local as the core component in this work but also confirms the role of GMW in feature selection through its enhancement over Non-Local. In the simple concatenation section, which replaces the proposed cascaded attention structure with basic concatenation, GMW-Non-Local still exhibits optimal angular error of 3.83°, demonstrating its stronger feature representation capability in gaze estimation. And this section's performance is worse than most results in the independent replacement section, further validating the effectiveness of our proposed cascaded attention structure. Consequently, the combined configuration section compares interactions between hybrid channel-spatial attention methods and GMW-Non-Local/Agent Attention to systematically evaluate the synergistic effects of different combination patterns. The results reveal that most combined attention mechanisms outperform their standalone counterparts, with the collaborative combination of CBAM and GMW-Non-Local achieving the best performance of 3.74°.

\subsection{Parameter Analysis}
In addition, we analyze the impact of the hyperparameter rounds $k$ for the cascaded attention structure in DMAGaze under the condition of $\sigma = 1.0$. As shown in Table~\ref{tab4}, the superior performance at $k=4$, with an angular error of 3.74°, indicates a balance between effective feature representation and generalization. When $k=2$, the limited iterations may result in insufficient feature extraction and information capture. In contrast, at $k=8$, the excessive iterations could lead to overfitting, thereby reducing the model's ability to generalize to unseen data. We also obtain comparable results for the parameter $\sigma$ when $k = 4$. By controlling the width of the Gaussian distribution, we get the lowest angular error at $\sigma = 1.0$, indicating that the balance between local and global information is optimal for capturing features relevant to gaze.

\subsection{Ablation Study}
We conduct ablation studies on the widely used MPIIFaceGaze gaze estimation dataset to explore the impact of different modules in DMAGaze. They are presented as follows: (1) \textbf{Baseline}: gaze estimation is performed using only the eye branch. (2) \textbf{Face branch (w/o MS-GLAM)}: the face branch is introduced to disentangling gaze-relevant and gaze-irrelevant features using the Disentangler, where "w/o MS-GLAM" indicates the absence of MS-GLAM. (3) \textbf{Head pose}: head pose features extracted using simple convolution are incorporated into the final gaze estimation. (4) \textbf{MS-GLAM (w/o GMW)}: the MS-GLAM is added, where "w/o GMW" indicates the exclusion of GMW. (5) \textbf{GMW}: the GMW is added to the Non-Local module in MS-GLAM, representing the full model.

\begin{figure}[t]
    \centering
    \includegraphics[width=0.6\columnwidth]{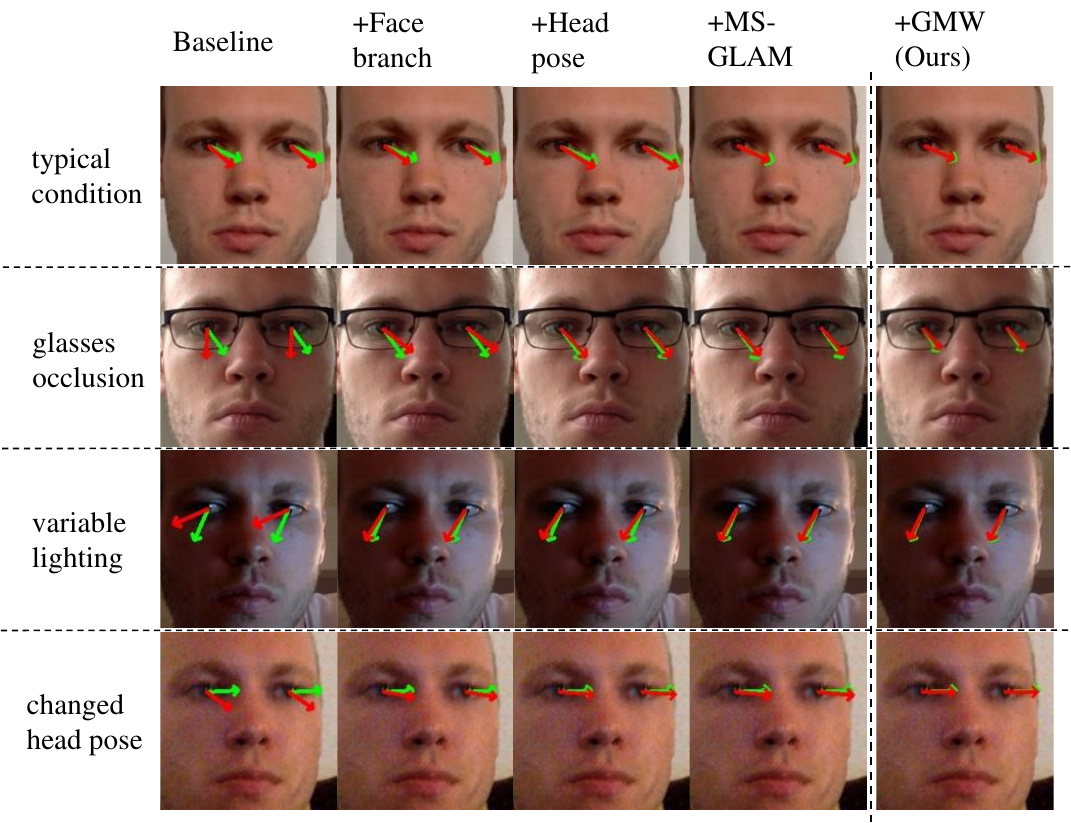} 
    \caption{The visualization of gaze estimation results in different components of our gaze estimation model from ablation studies, covering scenarios such as wearing glasses, variable lighting and head poses. The green line represents the ground truth gaze and the red line represents the estimated gaze.}
    \label{fig3}
\end{figure}

Table~\ref{tab3} presents the quantitative results of the ablation studies. The angular error of our baseline is 5.02°. Adding the face branch reduces the error by 19.52\%, partially validating the effectiveness of our cognitively inspired framework, which disentangles gaze-relevant facial information by reconstructing the eye regions and non-eye regions individually. Incorporating head pose features reduces the error by 4.46\%, proving its contribution to gaze estimation. Adding MS-GLAM (w/o GMW) achieves an error of 3.79°, demonstrating our cascaded attention structure's ability to integrate global and local information at multiple scales. Finally, introducing the GMW reduces the error by 1.32\%, proving the effectiveness of using Gaussian similarity to replace the dot product similarity computation in our framework. Fig.~\ref{fig3} shows the gaze estimation performance of each component in ablation studies under various conditions, such as glasses occlusion, variable lighting and head poses. For consistency, gaze directions are visualized with a standardized origin at the eye center. This results further validate the contributions of each module in our model.

\section{Conclusion}
We have presented a novel gaze estimation framework, DMAGaze, which addresses the complex relationships between facial and eye regions for gaze estimation, achieving accurate results. We introduced a Disentangler with a continuous mask matrix to disentangle gaze-relevant facial features. Then we designed a customized cascaded attention structure that integrates global and local information at multiple scales. Additionally, we introduced an innovative GMW by replacing the traditional dot product similarity calculation with a Gaussian similarity-based distance metric. The distilled global gaze-relevant features are ultimately integrated with local eye features and head pose features for final gaze estimation. Experiments demonstrate that DMAGaze achieves state-of-the-art performance on the well-known gaze estimation datasets with angular errors of 3.74° and 6.17°. Ablation studies further validate the contributions of individual modules. Our method exhibits excellent performance and robustness, highlighting the potential of this framework for applications in human-computer interaction and
other areas.

\bibliographystyle{elsarticle-num}   
\bibliography{references}
\end{document}